\documentclass[preprint,12pt,authoryear]{elsarticle}

\usepackage[utf8]{inputenc}
\usepackage{amsmath}
\usepackage{amsfonts}
\usepackage{amssymb}
\usepackage{latexsym}
\usepackage{graphicx}
\usepackage{hyperref}
\usepackage[OT4]{fontenc}
\usepackage{colortbl}

\let\tnote\relax
\usepackage{ctable}
\usepackage{soulutf8}

\usepackage{breakcites}
\usepackage{soul}
\usepackage{xcolor}
\usepackage{subcaption}
\definecolor{cb}{RGB}{223, 255, 128}
\definecolor{csb}{RGB}{159, 255, 128}
\definecolor{cw}{RGB}{128, 223, 255}
\definecolor{csw}{RGB}{128, 191, 255}

\let\R\relax
\let\set\relax
\newcommand{\R}{\mathbb{R}}
\newcommand{\set}[1]{\mathcal{#1}}
\newcommand{\vect}[1]{\boldsymbol{#1}}
\DeclareMathOperator{\Tr}{Tr}
\DeclareMathOperator{\Cov}{Cov}

\newcommand{\heuristic}[1]{\emph{#1}}

\begin{document}
\begin{frontmatter}
\title{Data structure $\geq$ labels? Unsupervised heuristics for SVM hyperparameter estimation}


\author[iitis]{Michał Cholewa}
\ead{mcholewa@iitis.pl}
\author[iitis]{Michał Romaszewski}
\author[iitis]{Przemysław Głomb}

%
%

\affiliation[iitis]{organization={Institute of Theoretical and Applied Informatics, Polish Academy of Sciences},
	addressline={Bałtycka~5}, 
	city={Gliwice},
	postcode={44100}, 
	country={Poland}}


\begin{abstract}
	Classification is one of the main areas of pattern recognition research, and within it, Support Vector Machine (SVM) is one of the most popular methods outside of field of deep learning -- and a de-facto reference for many Machine Learning approaches. Its performance is determined by parameter selection, which is usually achieved by a time-consuming grid search cross-validation procedure (GSCV). That method, however relies on the availability and quality of labelled examples and thus, when those are limited can be hindered. To address that problem, there exist several unsupervised heuristics that take advantage of the characteristics of the dataset for selecting parameters instead of using class label information. While an order of magnitude faster, they are scarcely used under the assumption that their results are significantly worse than those of grid search. To challenge that assumption we have proposed an improved heuristics for SVM parameter selection and tested it against GSCV and state of art heuristic on over 30 standard classification datasets. The results show not only its advantage over state-of-art heuristics but also that it is statistically no worse than GSCV.
\end{abstract}




\end{frontmatter}

\section{Introduction}

Classification is among the most frequently encountered problems within the field of pattern recognition. It is utilized, among many other fields, in computer vision \cite{koklu2020multiclass}, document analysis \cite{shah2020comparative}, data science \cite{alloghani2020systematic} and biometrics \cite{yang2021biometrics}. The classification itself is a wide area that contains both traditional machine learning methods and, recently increasingly popular, deep learning models. 
However, even with formidable results achieved by the deep learning approaches, e.g. \cite{li2021plant}, \cite{cheng2020remote}, the classical methods still have a role to play. The high computational cost, large data volume required and the open-ended difficulty of finding a combination of a suitable architecture, hyperparameters and a learning algorithm for the deep learning model is prohibitive for many current applications of pattern recognition. This situation occurs e.g. for Internet of Things devices~\cite{Menter2021IoT}, edge computing~\cite{Gupta2022Emails}, medical devices~\cite{Pires2018Limitations} or with limited training labels~\cite{romaszewski2016semi}. Additionally, classical methods -- notably Support Vector Machines -- are selected for their robustness~\cite{cervantes2020comprehensive} or theoretical consideration~\cite{huang2022provably}.

Support Vector Machine (SVM) is a supervised classification scheme based on ideas developed by V. N. Vapnik and A. Ya. Chervonenkis in 1960s~\cite{vapnik1963recognition} and later expanded on in works such as~\cite{smola1998learning}, \cite{cortes1995support} or~\cite{drucker1997support}. It is based on computing a hyperplane that optimally separates training examples and then making classification decisions based on the position of a point in relation to that hyperplane. The SVM have been consistently used in various roles -- as an independent classification scheme e.g. \cite{sha2020knn}, \cite{glomb2018application}, \cite{direito2017realistic}, part of more complex engines e.g. \cite{kim2003constructing}, \cite{cholewa2019spatial} or a detection engine e.g. \cite{ebrahimi2017detection}, \cite{chen2005application}. It has been also employed in unsupervised setting in works such as \cite{lecomte2011abnormal}, \cite{song2009unsupervised}. This flexibility allows SVM to be one of the most frequently used machine learning approaches in medicine \cite{subashini2009breast}, remote sensing \cite{romaszewski2016semi}, threat detection \cite{parveen2011supervised}, criminology \cite{wang2010predicting}, and is often utilized in photo, text, and time sequence analysis \cite{li2013hog}. In numerous studies, SVM is consistently marked as one of the top performing method~\cite{cervantes2020comprehensive}.

The popularity and versatility of SVM is to a large degree due to its controllability by the key hyperparameters. The first is a label error regularization coefficient $C$, which balances training error and margin width. It allows to classify non-linearly separable datasets or preserve margin width at the cost of misclassification of some training examples. The second is related to extension with the `kernel trick' to kernel-SVM, which is much more effective in working with complex data distributions; it introduces a kernel function value computation as an extension of a dot-product. Various kernel functions have been investigated, however, overwhelming majority of applications use Gaussian radial basis function as it provides best classification performances on a large range of datasets~\cite{fernandez2014we} and assumes only smoothness of the data, which makes it a natural choice when knowledge about data is limited~\cite{scholkopf1998regularization}. Values of these hyperparameters are typically found through supervised search procedures, cross-validation (CV) on the training set and grid-search through a range of predefined parameters~\cite{an2007fast} \cite{zhang2016fast}. 
However, major disadvantage of the CV is the $\mathcal{O}(n^2)$ complexity in the number of hyperparameter values to be evaluated, each requiring training a separate model. This is a burden for performing pattern recognition in distributed edge computing devices in Industry 4.0 \cite{Gupta2022Emails} or optimization of battery usage for mobile devices with limited connectivity, e.g. in monitoring of ageing people~\cite{Pires2018Limitations}.

An alternative for hyperparameter selection is to derive their values from a statistical analysis of the data. Those approaches range from simple `rule of thumb' statistics, e.g.~\cite{smola2011blog}, to more complex approaches involving e.g. cluster assumptions and graph distances between datapoints~\cite{chapelle2005semi}. Through those approaches, values of $C$ and $\gamma$ can be estimated based on structure of entire available dataset, in a unsupervised way -- without the requirement of labels. This is especially useful for applications that acquire large amount of data with limited supervision, e.g. IoT devices~\cite{Menter2021IoT}. Additionally, this estimation is one-pass computation much less intensive than cross-validation, allowing for much greater applicability, e.g. in IoT/edge/medical supervision devices. Unsupervised estimation avoids the issues of optimizing parameters on the same set as the one used for training, which can lead to overfitting~\cite{smola1998learning}. It is known that in some cases, e.g. where classes indeed do conform to the cluster assumption and Gaussian distribution~\cite{varewyck2010practical}, optimal or close to optimal parameter values can be analytically derived from data without knowledge about class labels. This approach is also very helpful when training data is very limited and may poorly reflect true class distributions -- a situation typically encountered in semi-supervised hyperspectral classification, e.g.~\cite{romaszewski2016semi}. The robustness of this approach has lead unsupervised heuristics to be a default parameter setting in SVM programming libraries, e.g. scikit-learn~\cite{gelbart2018github}.


In this work, we experimentally verify the performance of unsupervised heuristical hyperparameter estimation for an SVM classifier (UH-SVM) against a grid search CV trained SVM (GSCV-SVM). We evaluate a large set of unsupervised heuristics, and propose an extension aimed at improving performance of one of the most general approach -- Chapelle's heuristics~\cite{chapelle2005semi}. Our experiments show, that without specific prior knowledge of a dataset, there's a significantly higher chance of a number of UH-SVM approaches having similar or better accuracy than GSCV-SVM -- in terms of statistical significance of the results -- than to have a worse accuracy. Considering the significantly lesser requirement of computation power of UH-SVM with respect to GSCV-SVM, this in our opinion validates the conclusion of UH-SVM parameter estimation being in many application cases on par with the grid search. As part of results we show that our proposed extension of Chapelle's heuristics obtains results practically equivalent to GSCV. Additionally, while there are numerous works investigating individual heuristics, to the best of the authors' knowledge, there is no work that collects them together and compares them with each other.



\section{Methods}

In the following section, we will recall both the ideas behind the Support Vector Machines classifier and the heuristics that we include in our experiments. In some cases our unified presentation of them allows us to derive natural generalizations, e.g. a scaling of~\cite{chapelle2005semi} in high dimensional datasets or correction for~\cite{soares2004meta}.

\subsection{Kernel SVM}

A kernel SVM~\cite{smola1998learning} is a classifier based on the principle of mapping the examples from the input space into a high-dimensional feature space and then constructing a hyperplane in this feature space, with the maximum margin of separation between classes. Let $\set{X}\subset\R^{n}$ be a set of data and let $\mathbf{x}_i \in \set{X}, i=1, \ldots, m$ be the set of labelled examples. Let also $\mathcal{Y}=\{-1,1\}$ be a set of labels. We define a training set as a set of examples with labels assigned to them,
\begin{equation}
	\set{L}=\left\{(\mathbf{x}_i,y_i),i=1,\dots,m\right\}
	\quad
	\mathbf{x}_i\in\set{X}
	\quad
	y_i\in\set{Y}.
\end{equation}

The SVM assigns an example $\mathbf{x}\in\set{X}\subset\R^{n}$ into one of two classes using a decision function
\begin{equation}
	f(\mathbf{x}) = \mathrm{sgn}\left(\sum_{i=1}^{m}y_i\alpha_i K(\mathbf{x},\mathbf{x}_i)+b\right).
\end{equation}
Here, $\alpha_i \geq 0$ and $b$ are coefficients computed through Lagrangian optimization -- maximization of margin, or distance from hyperplane to classes' datapoints on the training set. Training examples $\mathbf{x}_i$ where the corresponding values of $\alpha_i\neq0$ are called support vectors (SV). Since SVM is inherently a binary classifier, for multi-class problems several classifiers are combined e.g. using one-against-one method~\cite{hsu2002comparison}.

\subsubsection{Kernel function}

The function $K:\set{X}\times\set{X}\rightarrow\R$ is called the kernel function and it is used to compute the similarity between the classified example $\mathbf{x}$ and each training instance $\mathbf{x}_i$. It is a generalization of a dot product operation used in the original linear SVM derivation,  i.e. $K(\mathbf{x},\mathbf{x}_i)=\langle\mathbf{x},\mathbf{x}_i \rangle$, taking advantage of the `kernel trick'~\cite{smola1998learning} -- a non-linear mapping $\phi:\set{X} \rightarrow \set{H}$ to a feature space $\set{H}$ where the dot product is computed by evaluating the value $K(\mathbf{x},\mathbf{x}_i)=\langle\phi(\mathbf{x}),\phi(\mathbf{x}_i)\rangle$. The kernel trick allows the SVM to be effectively applied in the case where classes are not linearly separable in the data space. A number of positive definite symmetric functions can be used as kernels, such as polynomial  $K(\mathbf{x},\mathbf{x}_i)=(\langle\mathbf{x},\mathbf{x}_i \rangle + c)^k$, $c\geq 0$, $k=1,2,\dots$; Laplace  $K(\mathbf{x},\mathbf{x}_i)=\exp\left(\frac{-||\mathbf{x}-\mathbf{x}_i||}{\sigma}\right)$ or Gaussian radial basis function (RBF):
\begin{equation}
	\label{eq:rbf_sigma}
	K(\mathbf{x}, \mathbf{x}_i) = \text{exp}\left(-\frac{\|\mathbf{x} - \mathbf{x}_i\|^2}{2 \sigma^2}\right),
\end{equation}
where $\sigma^2$ represents the variance of the data and $\|\cdot\|$ is an Euclidean distance in $\set{X} \subset \R^n$. This kernel has been found to be versatile and effective for many different kinds of data \cite{prajapati2010performing} and it will be the focus of our research. By substituting $\gamma = \frac{1}{2\sigma^2}$, it can be written:
\begin{equation}
	\label{eq:rbf_gamma}
	K(\mathbf{x}, \mathbf{x}_i) = \text{exp}\left(-\gamma \|\mathbf{x} - \mathbf{x}_i\|^2\right),
\end{equation}
where $\gamma$ can be viewed as scaling factor, which is one of the parameters of the SVM classifier.

The parameter $\gamma$ controls the impact of individual SV as the kernel distance between two examples decreases with higher values of $\gamma$. Therefore, small values of $\gamma$ will result in many SV influencing the point under test $\mathbf{x}$, producing smooth separating hyperplanes and simpler models. Very small values will lead to all SV having a comparable influence, making the classifier behave like a linear SVM. Large values of $\gamma$ result in more complex separating hyperplanes, better fitting the training data. However, a too high value of $\gamma$ may lead to overfitting (see Figure~\ref{fig:svmpar}).

\subsubsection{Soft margin}

\begin{figure}
	\begin{subfigure}[t]{.48\linewidth}
		\includegraphics[width=\textwidth]{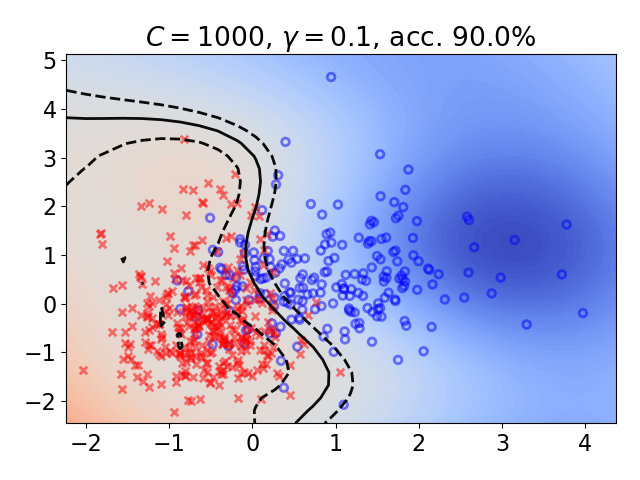}
		\caption{Good parameter values.}
	\end{subfigure}
	\begin{subfigure}[t]{.48\linewidth}
		\includegraphics[width=\textwidth]{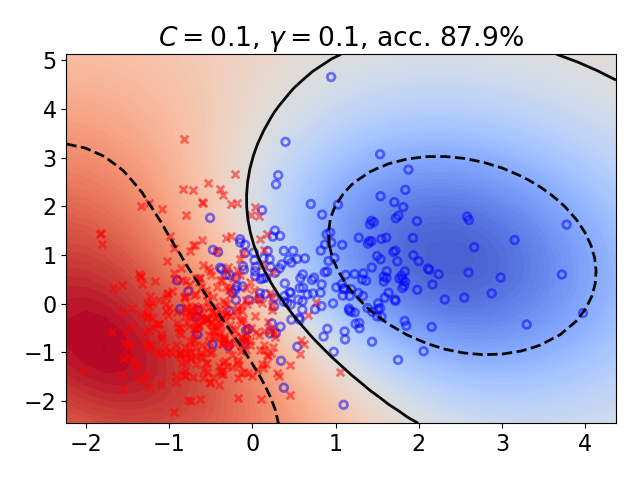}
		\caption{Lower $C$, larger margin.}
	\end{subfigure}
	\begin{subfigure}[t]{.48\linewidth}
		\includegraphics[width=\textwidth]{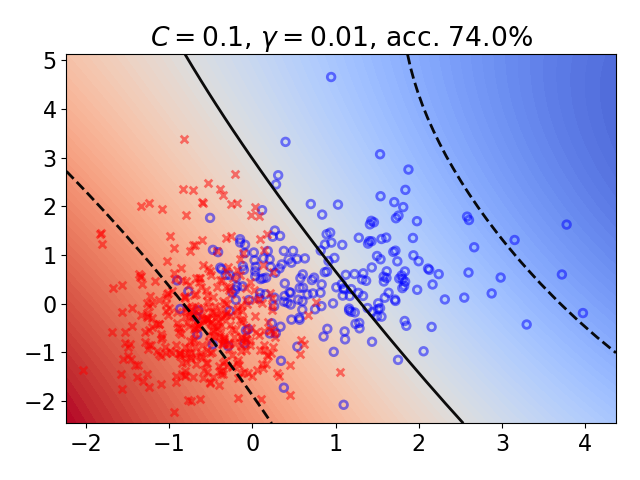}
		\caption{Lower $C$ and $\gamma$, larger margin, decision boundary `pushed away' from the more compact class.}
	\end{subfigure}
	\begin{subfigure}[t]{.48\linewidth}
		\includegraphics[width=\textwidth]{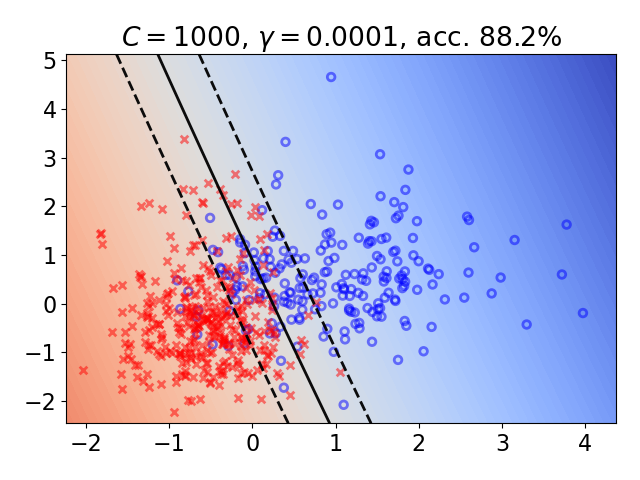}
		\caption{Very low $\gamma$, decision boundary approaching linear SVM.}
	\end{subfigure}
	\begin{subfigure}[t]{.48\linewidth}
		\includegraphics[width=\textwidth]{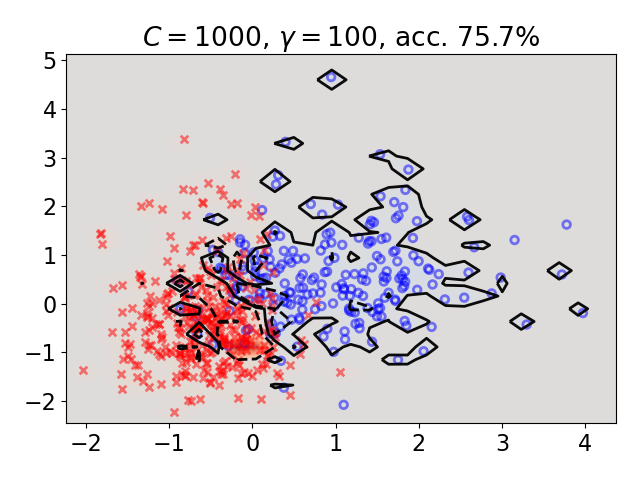}
		\caption{High $\gamma$, decision boundary approaching overfitting.}
	\end{subfigure}
	\begin{subfigure}[t]{.48\linewidth}
		\includegraphics[width=\textwidth]{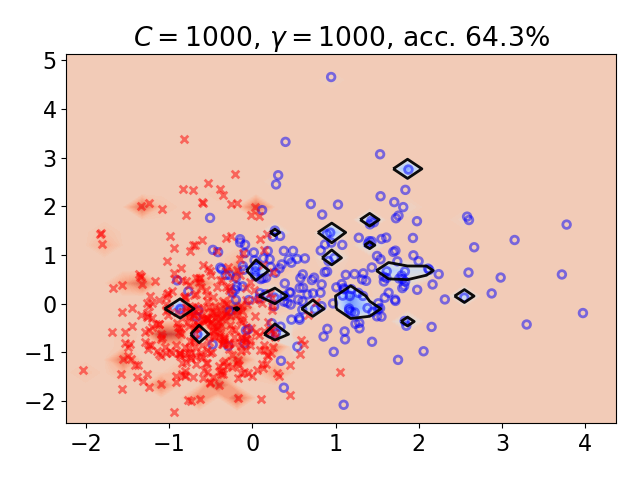}
		\caption{Very high $\gamma$, degenerate decision boundary.}
	\end{subfigure}
	\caption{Example SVM behaviour on first two features from the `Breast Cancer Wisconsin Dataset' (wdbc). Red crosses and blue circles mark the position of data points from two classes. Solid line presents the decision boundary, dashed lines denote margin ranges. Presented cases show the example influence of values of $C$ and $\gamma$ parameters, both for good and bad values.}
	\label{fig:svmpar}
\end{figure}

In practice, even using a kernel trick, a hyperplane that separates classes may not exist. Therefore, SVM is usually defined as a soft margin classifier by introducing slack variables to relax constraints of Lagrangian optimisation, which allows some examples to be misclassified. It introduces the soft margin parameter $C>0$ where $0<\alpha_i<C$ a constraint on $\alpha_i$ controlling the penalty on misclassified examples and determining the trade-off between margin maximization and training error minimization. Large values to the parame ter $C$ will result in small number of support vectors while lowering this parameter results in larger number of support vectors and wider margins (see Figure~\ref{fig:svmpar}).

\subsection{Setting the SVM parameters}

One of the early discussions about SVM parameters was provided in \cite{smola1998learning}. In the chapter~7.8, the authors mentioned the grid search CV (GSCV) as a common method of SVM parameter selection. As an alternative, in order to avoid the CV, the authors suggested a number of general approaches including scaling kernel parameters such as the denominator of the RBF kernel so that the kernel values are in the same range. They also suggested that the value of the parameters $C$ can be estimated as $C\propto 1/ R^2$ where $R$ is some measure of data variability such as standard deviation of the examples from their mean, or the maximum/average distance between examples.  Model selection by searching the kernel parameter space was later discussed in~\cite{chapelle2000model}, where authors proposed two simple heuristics based on leave-one-out CV. 

Unsupervised heuristics are relatively less discussed than their supervised counterparts. A simple heuristic that estimates $\gamma$ as an inverse of some aggregate (e.g. a median) of distances between data points has been proposed in a blog post~\cite{smola2011blog}. In fact, when searching the Internet for a method to choose kernel parameters in an unsupervised way, this post -- which refers to the idea from a thesis of B. Sch{\"o}lkopf 
-- is a common find. This heuristics is similar to the `sigest'\footnote{Implemented e.g. in R, see~ \cite{carchedi2021}} method~\cite{caputo2002appearance}. However, even 
in surveys comparing heuristics for SVM parameter selection~\cite{wainer2021tune} when sigest is considered it is applied to the training set and complimented with cross-validation for the value of the $C$ parameter. 

Sometimes, unsupervised heuristics supplement more complex methods, e.g. in~\cite{chapelle2005semi} authors propose a method for parameter selection inspired by the cluster assumption, based on graph distances between examples in the feature space; a heuristic 
for unsupervised initialisation of SVM parameters is provided as a starting point of a grid search. Another example are initialisation methods used in well-known ML libraries, e.g. scikit-learn\footnote{https://scikit-learn.org} employs its own implementation of heuristic for the $\gamma$ parameter~\cite{gelbart2018github} 
Shark\footnote{http://www.shark-ml.org/} uses the heuristic from~\cite{jaakkola1999using} and while this one is supervised, it can be used in an unsupervised way~\cite{soares2004meta}.

\subsubsection{Grid Search Cross Validation}
As a baseline method for model selection in this article, Grid Search Cross Validation (GSCV) \cite{berrar2019cross} is used. This method is based on dividing the dataset into $k$ parts $\{p_1, \ldots, p_k\}$ and then repeat the experiment using parts $\{p_1, \ldots, p_k\} \setminus \{p_i\}$ for training and $\{p_i\}$ for testing and averaging the results. This method allows to mitigate the variance resulting for random train/test set selection.

In case of this research, the additional layer is used for model selection -- called an internal layer. It is designed to detect the best set of parameters $(C, \gamma)$ from given grid $\set{G} \subset \mathbb{R}^2$.
Similarly to external layer, each training set $\set{T}_i = \{p_1, \ldots, p_k\} \setminus \{p_i\}$ is divided into $t$ subparts $\{p_i^1, \ldots, p_i^t\}$, with  $\{p_i^1, \ldots, p_i^t\} \setminus \{p_i^j\}$ used for training with given parameters from grid $\set{G}$ and $\{p_i^j\}$ used for testing (hence Grid Search Cross Validation). The parameters for  $\{p_i\}$ are determined by the results of this second level of cross validation.

\subsection{Unsupervised heuristics for $\gamma$}
\label{sec:unsupervised_heur}

Unsupervised heuristics usually assume that $\gamma$ should be relative to `average' distance (measured by $\|\cdot\|^2$) between the examples from $\set{X}$, so that the two extreme situations -- no SV influence or comparable influence of all SV -- are avoided. For example, $\gamma$ can be assigned the inverse of the data variance, which corresponds e.g. with heuristics described in~\cite{gelbart2018github} or \cite{smola2011blog}). Intuitively then kernel value between two points is a function of how large is the distance between two given points compared to the average distance among the data. Differences between heuristics can be thus reduced to different interpretations of what that average distance is.

\subsubsection{$\gamma$ heuristics for Gaussian-distributed data}
\label{sec:our_heuristics_gamma}
Considering a pair of examples $(\mathbf{x}_i,\mathbf{x}_j)\in \set{X}\times \set{X} \subset \R^n\times\R^n$ from Gaussian-distributed data, it has been noted in~\cite{varewyck2010practical}, that the squared Euclidean distance $\|\mathbf{x}_i - \mathbf{x}_j\|^2$ is Chi-squared distributed with a mean of $2n\sigma^2$, assuming that every data feature has variance $\sigma$ and mean $0$. This observation could be used as a heuristics to estimate the value of $\gamma$ as 
\begin{equation}
	\label{heu:varewyck}
	\gamma= \frac{1}{2n\sigma^2}.
\end{equation}
If we further assume that $\sigma^2=1$ , this simplifies to $\gamma=\frac{1}{2n}$, as noticed by authors of~\cite{wang2014super}.

This approach relies on an underlying assumption that data covariance matrix is in the form $\Cov(\vect{X})=\vect{I}\sigma^2$, which, in turn, means that in a matrix of examples $\vect{X}\in\R^{m\times n}$, every feature has an equal variance. In practice, data standardisation is used, which divides each feature by its standard deviation. However, the standard deviations are estimated on the training set, and on the test set will produce slightly varying values that are only approximately equal $\sigma_1 \approx \sigma_2 \approx \dots \approx \sigma_n$. To take that into account, we use another formula for estimation of the value of $\gamma$ as:
\begin{equation}
	\gamma = \frac{1}{2  \Tr\left(\Cov(\vect{X})\right)},
\end{equation} 
where $\Tr(\cdot)$ denotes a trace of a matrix. This heuristic is denoted in the experiments as \heuristic{covtrace}.

\subsubsection{Smola's  heuristics}\label{sub:Smola}
\label{sec:smola}
A well-known heuristics for computing the initial value of a parameter $\gamma$ was provided by A. J. Smola in an article on his website~\cite{smola2011blog}. Given examples $(\vect{x}_i,\vect{x}_j)\in\R^n\times\R^n$, he considered a kernel function in the form
\begin{equation}
	\label{eq:rbf_smola}
	K(\vect{x}_i, \vect{x}_j) = \kappa(\lambda \|\vect{x}_i - \vect{x}_j\|), 
\end{equation}
where a scaling factor $\lambda$ of this kernel is to be estimated and $\kappa : \R \rightarrow \R^+$. The Smola's kernel form is consistent with the RBF kernel given by Eq.~\eqref{eq:rbf_gamma} -- it as special case of \eqref{eq:rbf_smola}, with $\kappa(x) = \exp{(-x^2)}$ where $x\in\R$ and $\lambda = \sqrt{\gamma}$. 

He proposes to select a subset of (e.g. $m=1000$) available pairs $(\vect{x}_i,\vect{x}_j)$ and to compute their distances. Then, the value of  $\lambda$ can be estimated as the inverse of $q$ quantile (percentile) of distances where one of three candidates $q\in\{0.1, 0.5, 0.9\}$ is selected through cross-validation. The reasoning behind those values extends the concept of `average' distance: the value of $q=0.9$ corresponds to the high value of a scaling factor which results in decision boundary that is `close' to SV, $q=0.1$ corresponds to `far' decision boundary, $q=0.5$ aims to balance its distance as `average' decision boundary. The author argues that one of these values in likely to be correct i.e. result in an accurate classifier. Those three $q$ values are included in the experiments as \heuristic{Smola\_10}, \heuristic{Smola\_50} and \heuristic{Smola\_90}.

\subsubsection{Chapelle \& Zien $\gamma$ heuristics}
\label{sec:heu_chapelle}
A heuristic for choosing SVM parameters can be found in~\cite{chapelle2005semi}. Interestingly, to the best of our knowledge it is the only method that estimates both $C$ and $\gamma$ in an unsupervised setting (see \ref{heu:chapelle_C}). The heuristics take into account the density of examples in the data space. Authors introduce a generalization of a `connectivity' kernel, parametrized by $\rho>0$, which in the case of $\rho \rightarrow 0$ defaults to the Gaussian kernel. This kernel proposition is based on minimal $\rho$-path distance $D_{ij}^\rho$ which, for $\rho \rightarrow 0$ becomes Euclidean distance i.e. $D_{ij}^{\rho\rightarrow 0} = \|\vect{x}_i - \vect{x}_j\|_2$. 

Authors use the cluster assumption, by assuming that data points should be considered far from each other when they are positioned in different clusters. 
In~\cite{chapelle2005semi} authors consider three classifiers: Graph-based, TSVM and LDS. As this approach introduces additional parameters, which would make cross-validated estimation difficult, authors propose to estimate parameters through heuristics. 
The value of $\sigma$ (Equation \ref{eq:rbf_sigma}) is computed as  $\frac{1}{n_c}$-th quantile of  $\set{D}=\{D_{ij}^\rho: \vect{X}\times\vect{X}\in\R^n\times\R^n  \}$ where $n_c$ is the number of classes. For Gaussian RBF kernel this results in 
\begin{equation}
	\gamma=\frac{1}{2\text{ quantile}_{\frac{1}{n_c}}(\set{D})}
\end{equation}

Note that we consider only the case $\rho \rightarrow 0$, as only under this condition heuristics proposed in~\cite{chapelle2005semi} are comparable with other heuristics presented in this Section and compatible with our experiment. However, the authors' original formulation allows for other values of $\rho$. This heuristic, along with the complimentary for the $C$ parameter (see Section~\ref{heu:chapelle_C}) are denoted in the experiments as \heuristic{Chapelle}.

\subsubsection{Jaakkola's and Soares' heuristics}
\label{sec:jakkola} 

While the original Jaakkola's heuristics, described in \cite{jaakkola1999using} and \cite{jaakkola2000discriminative}, was supervised, in this article we will focus on its unsupervised version proposed in \cite{soares2004meta}.

The original heuristics based on median inter-class distance and is computed as follows: for all training examples  $\mathbf{x}\in\set{X}\subset\R^n$ we define $d_{min}^l(\mathbf{x})$ as a distance to its closest neighbour from a different class. Then a set of all nearest neighbour distances is computed as
\begin{equation}
	\set{D}^l = \left\{d_{min}^l(\mathbf{x}): \mathbf{x}\in\set{X}\right\},
\end{equation}
and the value of $\sigma = \text{median}(\set{D}^l)$.

This approach, however, has been interpreted differently in \cite{soares2004meta}, which resulted in an unsupervised heuristic based on what was proposed in \cite{jaakkola1999using}. The approach to estimate $\sigma$ is similar, however, it is calculated without any knowledge about labels of examples, which means that not inter-class but inter-vector distances are used. Considering an unlabelled distance $d_{min}(\mathbf{x})$ of an example $\mathbf{x}$ to its closest neighbour, the set of all neighbour distances is computed as
\begin{equation}
	\set{D} = \left\{d_{min}(\mathbf{x}): \mathbf{x}\in\set{X}\right\},
\end{equation}
and the value of $\sigma = \text{mean}(\set{D})$. This heuristic is denoted as \heuristic{Soares}.

The use of mean instead of median in an approach proposed in~\cite{soares2004meta} results in larger values of $\gamma$ in the case of outliers in the data space. Therefore, following the reasoning in the original manuscript~\cite{jaakkola1999using}, we propose to compute $\sigma = \text{median}(\set{D})$, which in case of the Gaussian RBF kernel results in:
\begin{equation}
	\gamma=\frac{1}{2 \text{ median}(\set{D})}.
\end{equation}
This heuristic is denoted as \heuristic{Soares\_med}.

\subsubsection{Gelbart's heuristics}
\label{sec:gelbart} 
The heuristic used to estimate the initial value of $\gamma$ in a well-known Python library scikit-learn, was proposed by Michael Gelbart in \cite{gelbart2018github}\footnote{\url{https://github.com/scikit-learn/scikit-learn/issues/12741}}. The scaling factor of Gaussian RBF kernel is computed as
\begin{equation}
	\gamma = \frac{1}{n \text{Var}(\set{X})},
\end{equation}
where $\set{X} \subset \R^n$  and $\text{Var}(\set{X})$ is a variance of all elements in the data set $\set{X}$.
It is easy to see that this heuristic is similar to the one discussed in Section~\ref{sec:our_heuristics_gamma}, based on~\cite{varewyck2010practical}: provided that every data feature has variance $\sigma$ and mean $0$ the value of Gelbart's heuristics is equal to the one described by Equation~\ref{heu:varewyck}. 
The advantage of this heuristics is its computational performance, and it has the potential to perform well when the variance of elements in the data array reflect the variance of the actual data vectors. This heuristic is denoted in our results as \heuristic{Gelbart}.

\subsection{Unsupervised heuristics for $C$}

Unsupervised heuristics for the $C$ parameter are much less common than for $\gamma$; in \cite{smola1998learning}, there is a suggestion that parameter $C \propto 1/R^2$, where $R$ is a measure for a range of the data in feature space and proposes examples of such $R$ as the standard deviation of the distance between points and their mean or radius of the smallest sphere containing the data. However, to the best of our knowledge, the only actual derivation of this idea was presented in~\cite{chapelle2005semi}, which we discuss below.

\subsubsection{Chapelle \& Zien $C$ heuristic}

\label{heu:chapelle_C}

Given a $\gamma$ value (originally computed as described in Section~\ref{sec:heu_chapelle}), \cite{chapelle2005semi} calculate the empirical variance:
\begin{equation}\label{eq:chapelle}
	s^2 = \frac{1}{m} \sum_{i=1}^{m}K(\mathbf{x}_i, \mathbf{x}_i) - \frac{1}{m^2} \sum_{i=1}^{m}\sum_{j=1}^{m} K(\mathbf{x}_i, \mathbf{x}_j),
\end{equation}
which, with $K(\vect{x}_i, \vect{x}_i)$ being the value of RBF kernel~\eqref{eq:rbf_gamma}, under the same $\rho \rightarrow 0$ assumption as Section~\ref{sec:heu_chapelle}, evaluates to
\begin{equation}
	s^2 = 1 - a,\qquad a=\frac{1}{m^2} \sum_{i=1}^{m}\sum_{j=1}^{m} K(\mathbf{x}_i, \mathbf{x}_j).
	\label{eq:chapelleC}
\end{equation}
The $C$ parameter value is then estimated as 
\begin{equation}
	C = \frac{1}{s^2}.
\end{equation}
This heuristic is denoted in our experiments as: \heuristic{Chapelle} when used in combination with authors' $\gamma$ heuristic (see Section~\ref{sec:heu_chapelle}) and \heuristic{+C} when used with \heuristic{covtrace} heuristic.

\subsubsection{A proposed extension of Chapelle \& Zien $C$ heuristic}

Our observations suggest that values of parameter $C$, when dealing with high-dimensional data such as hyperspectral images, should be higher than estimated with the heuristic proposed in Section~\ref{heu:chapelle_C}. Therefore we propose a new version of the heuristic, by modifying the formula~\ref{eq:chapelleC}. Since in formula~\ref{eq:chapelleC} the factor $a < 1$, higher values of $C$ can be achieved by substituting $s^2 = 1 - a'$ with $a \leq a' < 1$. 

The value of $a$ in Equation~\ref{eq:chapelleC} is an average of kernel values for all data points, which, for the RBF kernel, is a function of the average distances between the data points. By selecting a subset of the data points based on values of their distances, we can arbitrarily raise or lower the value of $a$. We start by considering a set of distances between the data points
\begin{equation}
	\mathcal{A} = \left\{ \| \mathbf{x}_i - \mathbf{x}_j\| : i, j \leq m; \:\mathbf{x}_i, \mathbf{x}_j \in \mathcal{X} \right\}.
\end{equation}
Then we define a subset of distances $\mathcal{A}'$ as $\frac{1}{n}$ quantile of $\mathcal{A}$ and we select a relevant set of data points pairs
\begin{equation}
	\mathcal{B} = \left\{(i,j) : \| \mathbf{x}_i - \mathbf{x}_j\| \in \mathcal{A}' \right\}.
\end{equation}
This leads to a modified version of the heuristic
\begin{equation}
	s^2 = 1 - a',\qquad a' = \frac{1}{t} \sum_{(i,j) \in \set{B}} K(\mathbf{x}_i, \mathbf{x}_j),
\end{equation}
with $t=\vert\mathcal{B}\vert$. The rationale of using $\frac{1}{n}$ quantile is that with increased dimension $n$, the proposed condition will restrict the set of pairs $\mathcal{B}$ to the distances between close points. This modified Chapelle's heuristic is denoted as \heuristic{+MC}, when used with \heuristic{covtrace} heuristic for $\gamma$.

\section{Experiments}
In this section we will present our method for experimental verification of unsupervised heuristics: the datasets that we use for tests, experimental procedure and finally our approach to statistical testing of obtained results.

\subsection{Datasets}
Experiments were performed using 31 standard classification datasets obtained from Keel-dataset repository~\footnote{https://sci2s.ugr.es/keel/category.php?cat=clas}, described in~\cite{alcala2011keel}.  Instances with missing values and features with zero-variance were removed, therefore the number of examples/features can differ from their version in the UCI~\cite{dua:2019} repository. The datasets were chosen to be diverse in regards to the number of features and classes and to include imbalanced cases. In addition, following~\cite{duch2012make}, the chosen set includes both complex cases where advanced ML models achieve an advantage over simple methods as well as datasets where most models perform similarly. Reference classification results can be found in\cite{moreno2012study} or through OpenML project~\cite{feurer2021openml}. The summary of the datasets used in experiments can be found in Table~\ref{tab:datasets}, along with the Overall Accuracy (OA) results of naive classifier (or zero-rule classifier, 0R) that classifies every point as the member of most frequent class. 

Before the experiment, every dataset was preprocessed by centering the data and scaling it to the unit variance. This operation was performed using mean and variance values estimated from the training part of the dataset.

\ctable[
cap     = Datasets,
caption = {Datasets used in the experiment. Balance is the ratio between size of the smallest and largest class. OA(0R) denotes the accuracy of a zero-rule, naive classifier that predicts the label of the most frequent class.},
label   = tab:datasets,
pos     = h,
doinside = {\tiny}]
{lrrrrrl}{\tnote[a]{As the dataset is named in KEEL repository~\url{https://sci2s.ugr.es/keel/datasets.php}}}
{\FL
	Name\tmark[a]&Examples&Features&Classes&Balance&OA(0R)&Notes or full name\ML
	appendicitis&106&7&2&0.25&80.2&\NN
	balance&625&4&3&0.17&46.1&Balance Scale DS\NN
	banana&5300&2&2&0.81&55.2&Balance Shape DS\NN
	bands&365&19&2&0.59&63.0&Cylinder Bands\NN
	cleveland&297&13&5&0.08&53.9&Heart Disease (Cleveland), multi-class\NN
	glass&214&9&6&0.12&35.5&Glass Identification\NN
	haberman&306&3&2&0.36&73.5&Haberman's Survival\NN
	hayes-roth&160&4&3&0.48&40.6&Hayes-Roth\NN
	heart&270&13&2&0.80&55.6&Statlog (Heart)\NN
	hepatitis&80&19&2&0.19&83.8&\NN
	ionosphere&351&33&2&0.56&64.1&\NN
	iris&150&4&3&1.00&33.3&Iris plants\NN
	led7digit&500&7&10&0.65&11.4&LED Display Domain\NN
	mammographic&830&5&2&0.94&51.4&Mammographic Mass\NN
	marketing&6876&13&9&0.40&18.3&\NN
	monk-2&432&6&2&0.89&52.8&MONK's Problem 2\NN
	movement-libras&360&90&15&1.00&6.7&Libras Movement\NN
	newthyroid&215&5&3&0.20&69.8&Thyroid Disease (New Thyroid)\NN
	page-blocks&5472&10&5&0.01&89.8&Page Blocks Classification\NN
	phoneme&5404&5&2&0.42&70.7&\NN
	pima&768&8&2&0.54&65.1&Pima Indians Diabetes\NN
	segment&2310&19&7&1.00&14.3&\NN
	sonar&208&60&2&0.87&53.4&Sonar, Mines vs. Rocks\NN
	spectfheart&267&44&2&0.26&79.4&SPECTF Heart\NN
	tae&151&5&3&0.94&34.4&Teaching Assistant Evaluation\NN
	vehicle&846&18&4&0.91&25.8&Vehicle Silhouettes\NN
	vowel&990&13&11&1.00&9.1&Connectionist Bench\NN
	wdbc&569&30&2&0.59&62.7&Breast Cancer Wisconsin (Diagnostic)\NN
	wine&178&13&3&0.68&39.9&\NN
	wisconsin&683&9&2&0.54&65.0&Breast Cancer Wisconsin (Original)\NN
	yeast&1484&8&10&0.01&31.2&\LL
}

\subsection{Choosing SVM parameters for a given dataset}
\label{svm_parameters}

The experiments used either one or two stages of cross-validation -- `external' and `internal' or `external' only -- depending on whether the grid search or heuristics were used. Let the heuristics $h\in\set{H}$ from the set of tested heuristics $\set{H}$ be a function that generates SVM parameters $\{C,\gamma\}$ based on a supplied training set $\set{T}$ i.e. $h:\set{T}\rightarrow\R^2$. We denote by $h_0$ a heuristic which always returns a pair $\{C,\gamma\}=\{1,1\}$, which are commonly assumed defaults, and thus a reference values which are not data-dependent. The $h_0$ heuristic is denoted in our experiments as \heuristic{default}.

For every training set $\set{T}_i$ corresponding with a given $i-$fold of the external CV, and for every heuristics $h\in\set{H}$ parameters of the SVM were selected in three ways:
\begin{enumerate}
	\item by performing a grid-search around the initial parameters $h_0$ and selecting the best model in the internal CV on $\set{T}_i$. 
	\item by applying the heuristics $h(\set{T}_i)$,
	\item by performing a grid-search around the initial parameters $h(\set{T}_i)$ and selecting the best model in the internal CV on $\set{T}_i$. 
\end{enumerate}

The range of parameters for GSCV to test is not always easy to determine as different studies propose different ranges - in \cite{matheny2007effects} the range $\{0, 0.1, 0.3, 0.5, 0.7\}$ is taken into consideration for $C$, while for $\gamma$ its $\{2^{-4}, 2^{-3}, \ldots 2^4\}$. Authors of \cite{schuhmann2021parameter} propose $C \in \{ x 10^y: x \in {1, 2, \ldots, 10}, y \in \{-2, -1, \ldots, 2\}\}, \gamma \in \{ x 10^y: x \in {1, 2, \ldots, 10}, y \in \{-4, -3, \ldots, 1\}\} $ while in research conducted in \cite{budiman2019svm} the selected range was $\{2^{-17}, 2^{-16}, \ldots 2^3\}$ for $\gamma$ and  $\{2^{-3}, 2^{-2}, \ldots 2^{17}\}$ for $C$. In \cite{lameski2015svm}, the authors decided to use the grid of ${10^{-6}, \ldots, 10^6}$ for both $C$ and $\gamma$. 

In this research, similar approach was selected, with range of parameters set as $\set{R}=\langle 10^{-5},10^{-4},..,10^{0},..,10^5\rangle$,  and the parameter grid $\set{G}_h$ for the heuristics $h$  generated as
\begin{equation}
	\set{G}_h = \left\{ r i_\gamma : r \in \set{R} \right\} \times \left\{ r i_C : r \in \set{R} \right\},
\end{equation}
where $h(\set{T}) = (i_\gamma, i_C)$. For the external CV, the number of folds $k_{\text{external}}=5$, for the internal CV the number of folds $k_{\text{internal}}=3$; both were stratified CVs, by which we mean the approach often used towards unbalanced sets which selects training and test sets maintaining similar percentage of datapoints from each class\footnote{we used implementation provided by https://scikit-learn.org/}.

For assessing classification performance, the Balanced Accuracy measure~\cite{brodersen2010balanced} (BA) was employed. BA can be expressed as the mean of classification accuracies in classes i.e. the mean between a ratio of correctly classified examples to the total number of examples in every class. Compared to the Overall Accuracy (OA), which is the ratio between a number of correctly classified examples to the total number of examples in dataset, it less sensitive to unbalance in class size. 

The final performance of the classifier in an experiment is the mean BA between external folds. Every experiment was repeated 10 times and the final values of BA were obtained by averaging the performance values of individual runs.

\subsection{Statistical verification of results}
\label{sec:statistical_verifictaion}
A typical approach to verify statistical significance of results is to use null hypothesis significance testing (NHST). While common, the NHST has several disadvantages explained in detail in~\cite{benavoli2017time}. Two particular ones are: the fact that point-wise null hypotheses are usually false, provided that sufficiently large number of data points is available, as in practice no two classifiers have perfectly similar accuracy; NHST does not allow to reach conclusion when the null hypothesis is rejected, which limits its usefulness. As an alternative, authors of~\cite{benavoli2017time} propose a new methodology based on Bayesian analysis that was adapted for analysing our results.  This methodology compares classifiers by estimating and querying the posterior distribution of their mean difference. The methodology introduces the \emph{region of practical equivalence} (rope) which refers to the value of mean difference that implies that classifiers are  practically equivalent e.g. when their accuracies differ by less then 1\%. This allows to infer the probability $P(\text{classifier}_A<\text{classifier}_B)$ of the mean difference between classifiers being practically negative which implies that classifier$_B$ is more accurate, as well as the probability of the opposite inequality and the probability $P(\text{classifier}_A=\text{classifier}_B)$ that both classifiers are practically equivalent with regards to the rope value. In addition the methodology allows for drawing conclusions through the simultaneous analysis of multiple data sets and it has a dedicated, clear visualisation of test results. 

Since we perform experiments using multiple datasets, the approach employing hierarchical models, described in Section~4.3.1 of \cite{benavoli2017time} was employed. Following the suggestion in~\cite{benavoli2017time}, the value of rope was set to 1\%. 




\subsection{Implementation}
SVM implementation was from the scikit-learn library v1.0.2. Bayesian comparison of classifiers~\cite{benavoli2017time} and its visualisation was performed using baycomp library v. 1.0.2\footnote{https://github.com/janezd/baycomp}. Matplotlib and seaborn libraries were used for data visualisation.

\section{Results and discussion}

\begin{figure}
	\centering
	\begin{subfigure}[b]{0.49\linewidth}
		\includegraphics[width=1.0\linewidth]{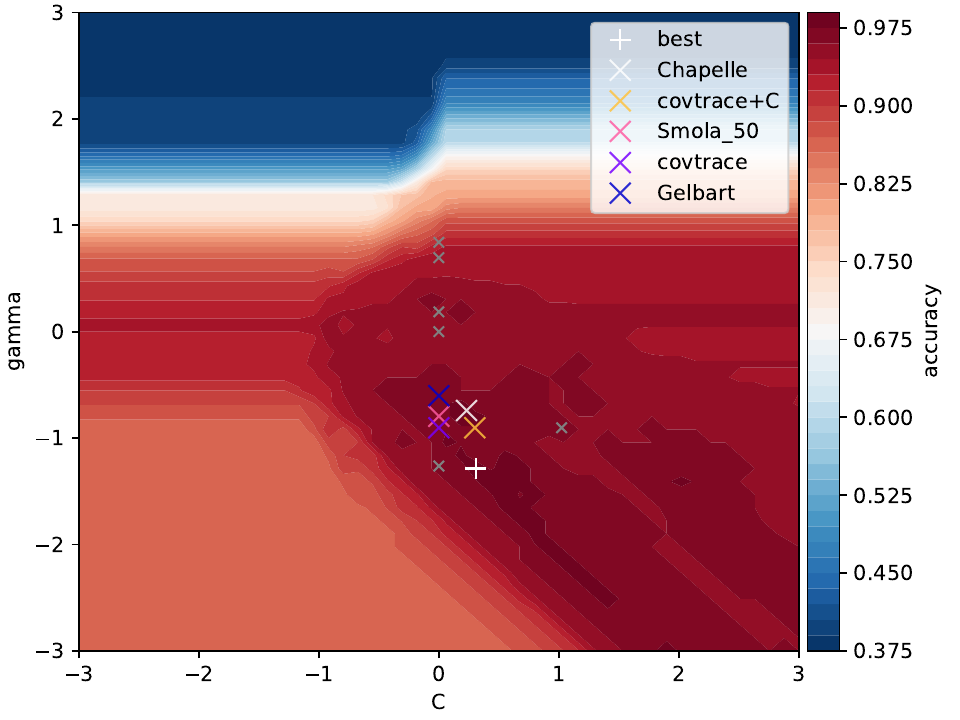}
		\caption{Iris}
	\end{subfigure}
	\begin{subfigure}[b]{0.49\linewidth}
		\includegraphics[width=1.0\linewidth]{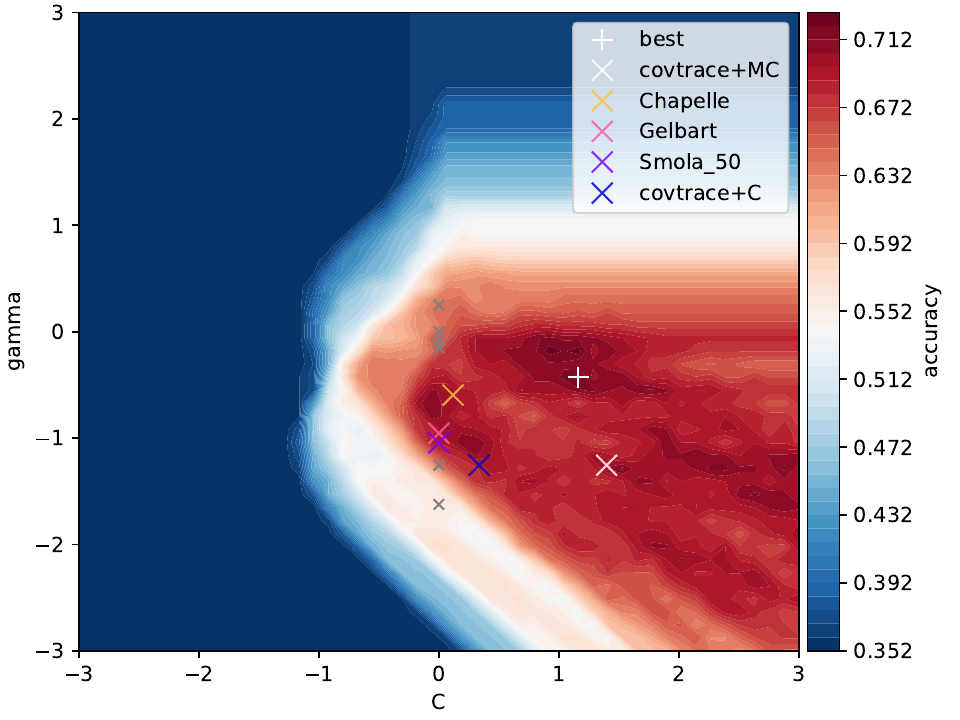}
		\caption{Glass}	
	\end{subfigure}
	\begin{subfigure}[b]{0.49\linewidth}
		\includegraphics[width=1.0\linewidth]{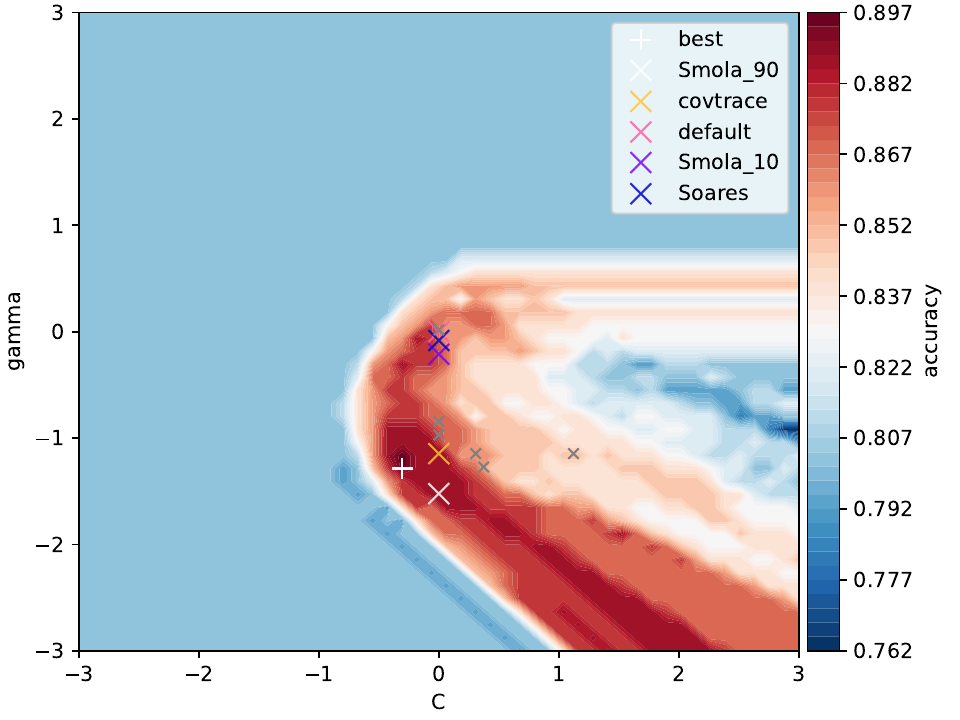}
		\caption{Appendicitis}
	\end{subfigure}
	\begin{subfigure}[b]{0.49\linewidth}
		\includegraphics[width=1.0\linewidth]{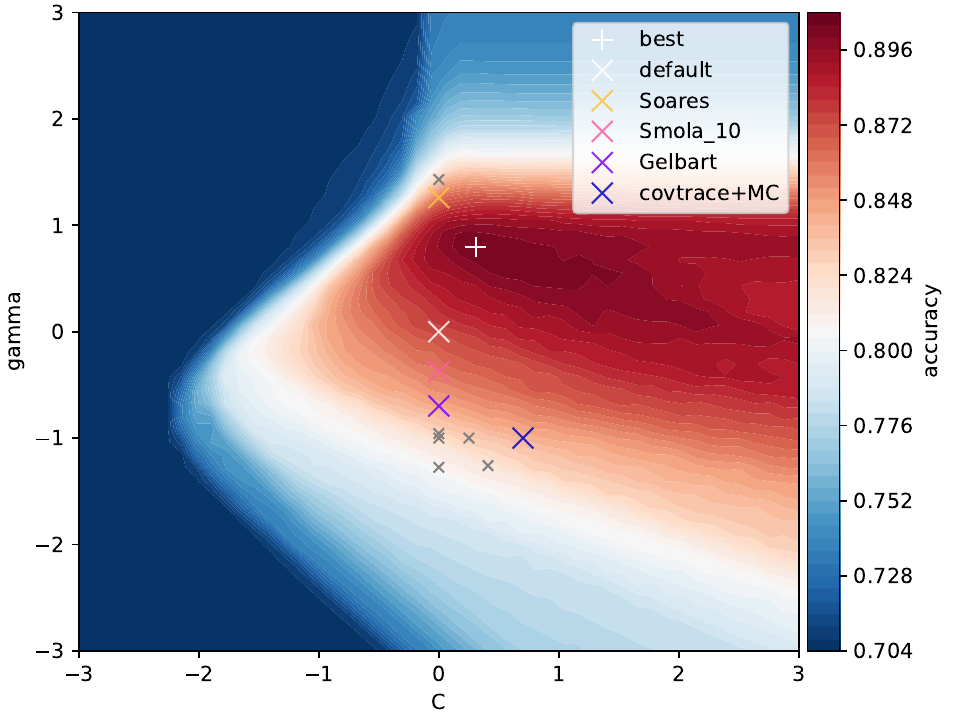}
		\caption{Phoneme}
	\end{subfigure}
	\begin{subfigure}[b]{0.49\linewidth}
		\includegraphics[width=1.0\linewidth]{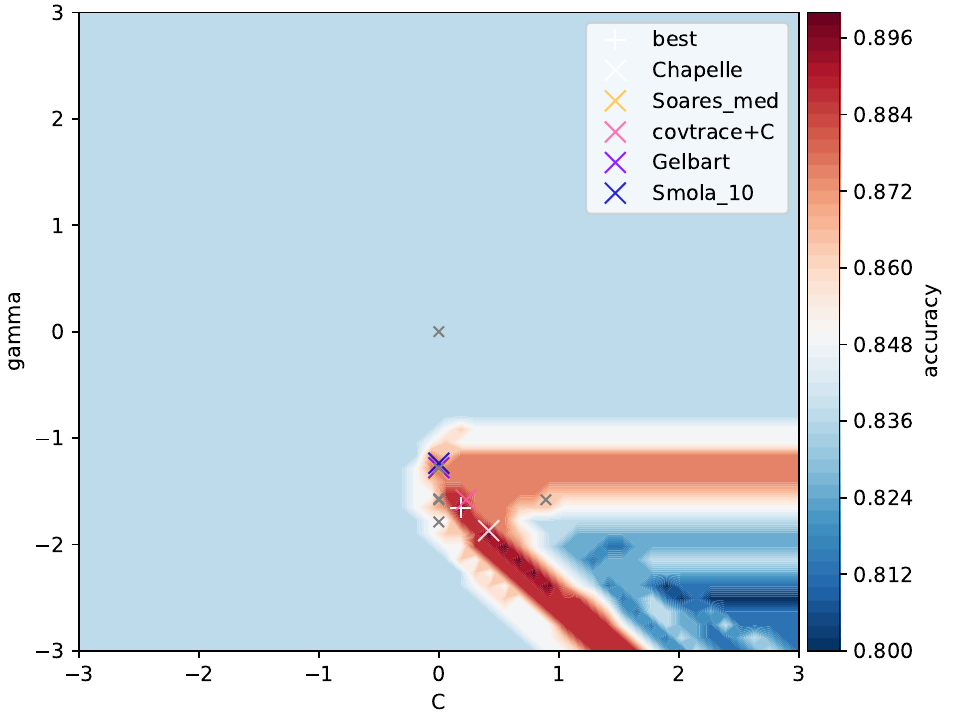}
		\caption{Hepatitis}
	\end{subfigure}
	\begin{subfigure}[b]{0.49\linewidth}
		\includegraphics[width=1.0\linewidth]{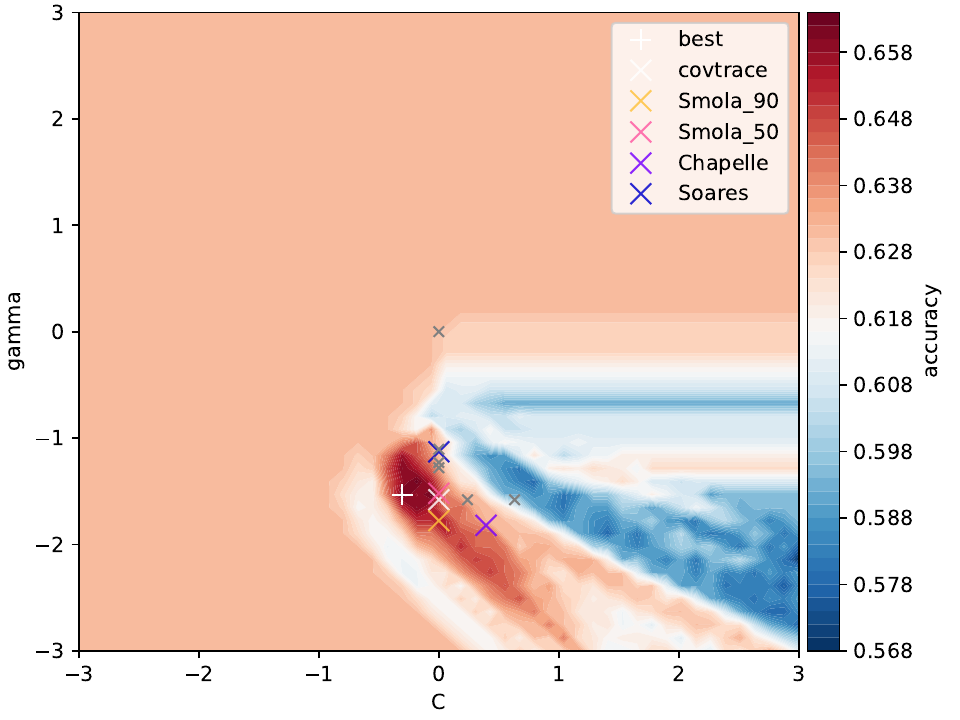}
		\caption{Bands}
	\end{subfigure}
	\caption{The impact of SVM parameters on its accuracy. Parameter values are presented in logspace. Accuracy values were obtained from experiments with 5-fold CV by sampling each pair of parameters from the $50\times50$ parameter grid. The highest value of accuracy is denoted as `best'. Marked points denote results of unsupervised heuristics from this paper, with the five heuristics scoring highest marked with colour.}
	\label{fig:cvresultsappendicitis}
\end{figure}

Our experiments compared the accuracy of the previously discussed UH-SVM approaches, to the GSCV-SVM, on the 31 Keel datasets. For each approach, the individual scores were aggregated into an estimated probability of practical advantage/disadvantage/equivalence of the heuristics and GSCV with regards to classifier accuracy. The summary of results for the Balanced Accuracy (BA) measure\footnote{For the reference, results of experiments for OA measure are presented in the Appendix~\ref{tab:results_bayes}.} is presented in Table~\ref{tab:results_bayes_aa}. Since most of the heuristics only estimate the $\gamma$ parameter, and only two of them estimate the $C$ (\emph{Chapelle}, \emph{MC}), we present results as a combination of every $\gamma$ and $C$ heuristics including the `default' value of $\gamma=1$, $C=1$. The advantage or any disadvantage of any one method corresponds to a sufficiently large difference between means of accuracies over all datasets, as described in Section~\ref{sec:statistical_verifictaion}; their practical equivalence corresponds to sufficiently small difference, with regards to rope value of 1\%. 

\begin{figure}
	\centering
	\begin{subfigure}[b]{0.49\linewidth}
		\includegraphics[width=1.0\linewidth]{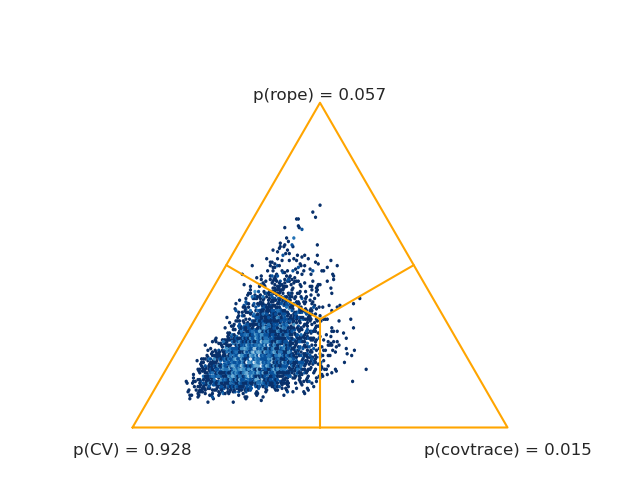}
		\caption{Covtrace, BA}
	\end{subfigure}
	\begin{subfigure}[b]{0.49\linewidth}
		\includegraphics[width=1.0\linewidth]{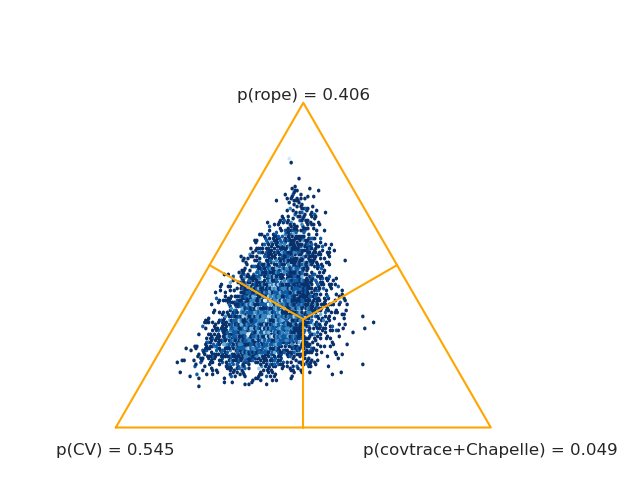}
		\caption{Covtrace+Chapelle, BA}	
	\end{subfigure}
	\begin{subfigure}[b]{0.49\linewidth}
		\includegraphics[width=1.0\linewidth]{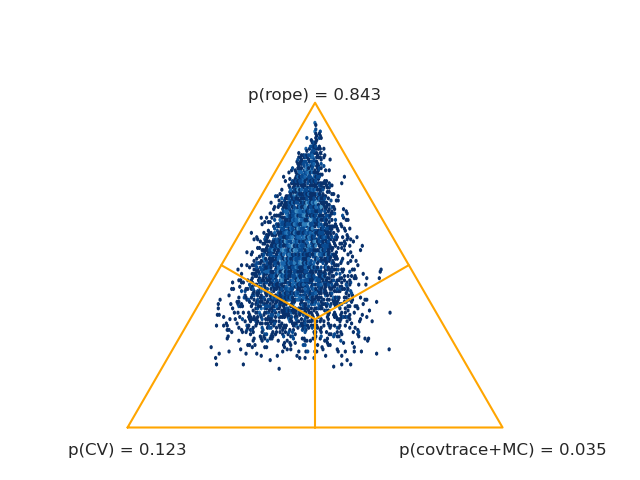}
		\caption{Covtrace+MC, BA}	
		\label{fig:bayesian_simplex_c}
	\end{subfigure}
	\begin{subfigure}[b]{0.49\linewidth}
		\includegraphics[width=1.0\linewidth]{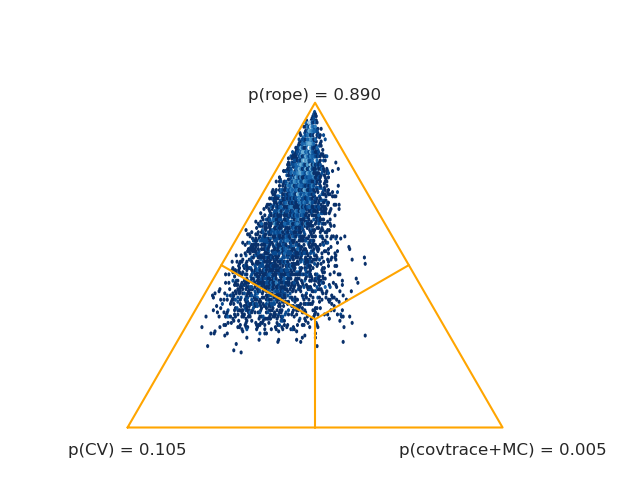}
		\caption{Covtrace+MC, OA}	
	\end{subfigure}
	\caption{Visualisation of Bayesian analysis of results with methodology from~\cite{benavoli2017time} for selected cases from Table~\ref{tab:results_bayes_aa}: \emph{covtrace+default}, \emph{covtrace+Chapelle}, \emph{covtrace+MC}. Vertices of the simplex represent decisions with certainty in favour of: CV (lower left), example heuristics (lower right) and rope (top); the latter corresponds to practical equivalence of CV and heuristics accuracy. Points represent Monte Carlo sampling of posterior probabilities in barycentric coordinates.  BA denotes balanced accuracy, OA denotes overall accuracy. Note that the better the $C$ heuristics, the closer to equivalence of UH-SVM and GSCV-SVM. Our proposed extension (\emph{MC}) provides the best results. The tendency visible in this plot is similar across other well-performing $\gamma$ heuristics.}
	\label{fig:bayesian_simplex}
\end{figure}

When no heuristics or only $\gamma$ heuristics are used, parameters obtained by GSCV result in significantly higher accuracy. There's only a marginal improvement when \emph{Chapelle} heusitics is used for selecting a $C$ parameter value. However, when using our proposed extension, the \emph{MC} heuristics, five of the $\gamma$ heuristics tested obtained the accuracy very close, or practically equivalent to CV. The combination of \emph{Covtrace+MC} resulted in the highest estimated value of this probability which indicates, that on average this heuristics results in classification accuracy no worse than GSCV. Visualisation of results for example heuristics is presented in~Figure~\ref{fig:bayesian_simplex}. The improvement in accuracy arising from the use of the two heuristics (three, if including the default) for the $C$ parameter is clearly evident in plots (a--c). Notably, the more effective the heuristic, the more equivalent are the scores of UH-SVM and GSCV-SVM.
Plot (d) presents similar results for an overall accuracy (OA) measure compared to the BA in plot (c). The use of OA measure usually results in slightly higher probabilities of practical equivalence between heuristics and GSCV. This suggests the class imbalance negatively affects GSCV's performance. 

The practical equivalence in the accuracy of classifiers whose parameters were chosen by GSCV and heuristics, is also visible during the inspection of the parameter values obtained from heuristics plotted on the graph showing the relationship between the classifier's effectiveness and its parameters (estimated through a dense grid of parameters). In the selected representative examples on Figure~\ref{fig:cvresultsappendicitis}, it can be seen that most of these points, especially for the best heuristics, are usually located in areas of high accuracy.

Interestingly, out of \emph{Smola} heuristics, the result of \emph{Smola$_{50}$+MC} resulted in the BA value most equivalent to GSCV. This indicates that the median distance between examples in the data space is of particular importance when choosing the $\gamma$ parameter.

Comparison of execution time for heuristics and GSCV is presented in Table~\ref{tab:time}. The values express a ratio of mean computation time of an experiment with GSCV parameter selection to experiment with parameters selected with heuristics. The average time was calculated over ten iterations of the experiment across all datasets. The use of heuristics allows, on average, to speed up calculations 100--200 times. Differences in times result not only from calculating the parameter values, but also from the impact of these values on the classifier -- increasing the value of the $\gamma$ and $C$ parameter extends the calculation time.

\ctable[
caption = {Results of experiments -- performance of different UH-SVM approaches with respect to GSCV-SVM. The numbers correspond to probabilities computed with the Bayesian analysis with methodology from~\cite{benavoli2017time}. Three right columns present probabilities of cross-validation being on average more/ equivalently / less accurate than heuristics. Results were obtained for the balanced accuracy measure and rope value of 1\%. Note that, with proposed (\emph{MC}) heuristic, several of $\gamma$ heuristics achieve results very close to GSCV.},
label   = tab:results_bayes_aa,
pos     = !h]
{lllll}{}
{\FL
C heuristics&$\gamma$ heuristics&P($\text{CV}>\text{H}$)&P($\text{CV}=\text{H}$)&P($\text{CV}<\text{H}$)\ML
 &default &  1.00 &    0.00 &   0.00 \\
 &Gelbart &  0.85 &    0.13 &   0.02 \NN
 &Smola\_10 &  0.97 &    0.02 &   0.01 \NN
 &Smola\_50 &  0.86 &    0.11 &   0.03 \NN
default &Smola\_90 &  0.99 &    0.00 &   0.01 \NN
 &Soares &  1.00 &    0.00 &   0.00 \NN
 &Soares\_med &  1.00 &    0.00 &   0.00 \NN
 &Chapelle &  0.98 &    0.01 &   0.01 \NN
 &covtrace &  0.93 &    0.06 &   0.01 \ML
& default &  1.00 &    0.00 &   0.00 \NN
&Gelbart &  0.60 &    0.36 &   0.04 \NN
&Smola\_10 &  0.96 &    0.03 &   0.01 \NN
&Smola\_50 &  0.68 &    0.24 &   0.08 \NN
Chapelle&Smola\_90 &  0.84 &    0.12 &   0.04 \NN
&Soares &  0.99 &    0.00 &   0.01 \NN
&Soares\_med &  0.99 &    0.00 &   0.01 \NN
&Chapelle &  0.83 &    0.14 &   0.03 \NN
&covtrace &  0.55 &    0.41 &   0.05 \ML
	&default &  1.00 &    0.00 &   0.00 \NN
	&Gelbart &  0.19 &    0.76 &   0.05 \NN
	&Smola\_10 &  0.68 &    0.28 &   0.04 \NN
	&Smola\_50 &  0.17 &    0.81 &   0.02 \NN
	MC&Smola\_90 &  0.34 &    0.60 &   0.07 \NN
	&Soares &  0.98 &    0.00 &   0.02 \NN
	&Soares\_med &  0.99 &    0.00 &   0.01 \NN
	&Chapelle &  0.21 &    0.76 &   0.02 \NN
	&covtrace &  0.12 &    0.84 &   0.03 \NN
}

\ctable[
caption = {Performance of heuristics as ratio of CV/heuristics execution time i.e. how many times heuristics is faster than CV. Times were estimated from 10 experiments and averaged over all datasets. Note that in almost all cases the speedup is 100--200 times.},
label   = tab:time,
pos     = h]
{llll}{}
{\FL
	&\multicolumn{3}{c}{C heuristics}\NN
	$\gamma$ heuristics&Default&Chapelle&MC\ML
	default &  136.88&    106.52 &   97.91 \\
	Gelbart &  248.72 &    182.27 &   149.33 \NN
	Smola\_10 &  153.92 &    136.47 &   121.71 \NN
	Smola\_50 &  169.13 &    149.29 &   131.29 \NN
	Smola\_90 &  158.75 &    149.64 &   131.19 \NN
	Soares &  132.01 &    105.50 &   94.59 \NN
	Soares\_med &  125.46 &    101.45 &   92.51 \NN
	Chapelle &  163.59 &    148.40 &   132.23 \NN
	covtrace &  237.38 &    188.61 &   154.86 \ML
}

To summarise: estimation of both parameters, in particular with \emph{Covtrace+MC} heuristics, leads to accuracy practically equivalent to GSCV (see Figure~\ref{fig:bayesian_simplex_c}) with parameters obtained in only $\sim$0.006 of its working time (see Table~\ref{tab:time}). 
Unsupervised heuristics for SVM parameters are likely effective because the test datasets conform to the clustering assumption, where data space forms structures/clusters useful to the classification problem and data point distributions reflect class divisions. However, the same assumption is the basis of training set selection with GSCV. As datasets deviate from the clustering assumption, the effectiveness of both approaches decreases, especially when training data is limited.

GSCV is by no means inferior to the heuristics, especially if supplied with proper number of labelled datapoints. In practice, however, the differences are often very small. and while it is natural to use GSCV when standard approach is preferable (small number of examples, training time is not an issue), in many scenarios i.e. processing on edge IoT devices, proposed heuristics offers practically equivalent accuracy in fraction of time..

\section{Conclusions}

In this study we evaluated unsupervised heuristics for SVM parameter selection on over thirty benchmark datasets, comparing their performance with GSCV. We have also proposed a modification to Chapelle \& Zien heuristics for the $C$ parameter as optimisation of both parameters is vital for accurate classifiers. We compared results with methodology based on Bayesian analysis, described in~~\cite{benavoli2017time}. Our results indicate that heuristics and in particular the proposed \emph{covtrace+MC}, are usually practically equivalent to GSCV i.e. obtained accuracies differ by less than 1\% (see Figure~\ref{fig:bayesian_simplex_c} and probabilities of equivalence in Table~\ref{tab:results_bayes_aa}). Moreover, these heuristics offer a computation time reduction, achieving a 100-200 times speed-up (see Table~\ref{tab:time}). This makes unsupervised, heuristic approach to parameters selection a compelling alternative for GSCV for rapid SVM calibration.

\ctable[
caption = {Results of experiments -- performance of different UH-SVM approaches with respect to GSCV-SVM, for Overall Accuracy (a supplement to Table~\ref{tab:results_bayes_aa}) The numbers correspond to probabilities computed with the Bayesian analysis with methodology from~\cite{benavoli2017time}. Three right columns present probabilities of cross-validation being on average more/ equivalently / less accurate than heuristics. Results were obtained for the rope value of 1\%. Note that, with proposed (\emph{MC}) heuristic, several of $\gamma$ heuristics achieve results very close to GSCV.},
label   = tab:results_bayes,
pos     = h]
{lllll}{}
{\FL
	C heuristics&$\gamma$ heuristics&P($\text{CV}>\text{H}$)&P($\text{CV}=\text{H}$)&P($\text{CV}<\text{H}$)\ML
	&default &  1.00 &    0.00 &   0.00 \NN
	&Gelbart &  0.70 &    0.25 &   0.05 \NN
	&Smola\_10 &  0.94 &    0.05 &   0.01 \NN
	&Smola\_50 &  0.61 &    0.35 &   0.04 \NN
	Default &Smola\_90 &  0.95 &    0.03 &   0.02 \NN
	&Soares &  0.99 &    0.00 &   0.00 \NN
	&Soares\_med &  1.00 &    0.00 &   0.00 \NN
	&Chapelle &  0.94 &    0.04 &   0.02 \NN
	&covtrace &  0.74 &    0.23 &   0.04 \ML
	&default &  1.00 &    0.00 &   0.00 \NN
	&Gelbart &  0.56 &    0.40 &   0.04 \NN
	&Smola\_10 &  0.92 &    0.07 &   0.01 \NN
	&Smola\_50 &  0.66 &    0.27 &   0.07 \NN
	Chapelle &Smola\_90 &  0.85 &    0.10 &   0.05 \NN
	&Soares &  1.00 &    0.00 &   0.00 \NN
	&Soares\_med &  1.00 &    0.00 &   0.00 \NN
	&Chapelle &  0.74 &    0.21 &   0.04 \NN
	&covtrace &  0.68 &    0.24 &   0.08 \ML
	&default &  1.00 &    0.00 &   0.00 \NN
	&Gelbart &  0.10 &    0.90 &   0.01 \NN
	&Smola\_10 &  0.54 &    0.44 &   0.01 \NN
	&Smola\_50 &  0.13 &    0.86 &   0.00 \NN
	MC &Smola\_90 &  0.41 &    0.58 &   0.02 \NN
	&Soares &  0.99 &    0.00 &   0.00 \NN
	&Soares\_med &  1.00 &    0.00 &   0.00 \NN
	&Chapelle &  0.19 &    0.80 &   0.01 \NN
	&covtrace &  0.10 &    0.89 &   0.01 \LL
}

\bibliographystyle{elsarticle-harv} 
\bibliography{heuristics_for_SVM}

\end{document}